\documentclass[10pt,twocolumn,letterpaper]{article}

\usepackage[pagenumbers]{cvpr} 

\usepackage{placeins}
\definecolor{cvprblue}{rgb}{0.21,0.49,0.74}
\usepackage[pagebackref,breaklinks,colorlinks,allcolors=cvprblue]{hyperref}

\def\confName{arXiv}
\def\confYear{2025}

\title{InstaFace: Identity-Preserving Facial Editing with Single Image Inference}

\author{
MD Wahiduzzaman Khan$^{1}$ \quad Mingshan Jia$^{1}$ \quad Xiaolin Zhang$^{2*}$ \quad En Yu$^{1}$ \\ Caifeng Shan$^{2}$ \quad Kaska Musial-Gabrys $^{1}$ \\[4pt]
$^1$University of Technology Sydney, Australia \\ 
$^2$Shandong University of Science and Technology, China \\[4pt]
\texttt{arnobk511@gmail.com} \quad
\texttt{mingshan.jia@uts.edu.au} \quad
\texttt{solli.zhang@gmail.com} \\ 
\texttt{en.yu-1@uts.edu.au} \quad
\texttt{caifeng.shan@gmail.com} \quad
\texttt{musial.katarzyna@gmail.com}
}

\begin{document}
\maketitle

\renewcommand\thefootnote{}
\footnotetext{$^*$Corresponding author}

\begin{abstract}

Facial appearance editing is crucial for digital avatars, AR/VR, and personalized content creation, driving realistic user experiences. However, preserving identity with generative models is challenging, especially in scenarios with limited data availability. Traditional methods often require multiple images and still struggle with unnatural face shifts, inconsistent hair alignment, or excessive smoothing effects. To overcome these challenges, we introduce a novel diffusion-based framework, InstaFace, to generate realistic images while preserving identity using only a single image. Central to InstaFace, we introduce an efficient guidance network that harnesses 3D perspectives by integrating multiple 3DMM-based conditionals without introducing additional trainable parameters. Moreover, to ensure maximum identity retention as well as preservation of background, hair, and other contextual features like accessories, we introduce a novel module that utilizes feature embeddings from a facial recognition model and a pre-trained vision-language model. Quantitative evaluations demonstrate that our method outperforms several state-of-the-art approaches in terms of identity preservation, photorealism, and effective control of pose, expression, and lighting.




\end{abstract}

\section{Introduction}

With the advancements in generative models ~\cite{goodfellow2014generative}, high-quality image synthesis has become widespread, significantly transforming the landscape of image editing ~\cite{kawar2023imagic, mokady2023null, roich2022pivotal} and reducing the reliance on manual operations in specialized applications. There have been notable successes in semantic-level tasks, such as converting an image into various artistic styles, \eg, anime, cinematic, retro, sketch, and altering objects in an image ~\cite{suvorov2022resolution, yu2018generative, zhang2019deep, chen2018deep}. However, it remains challenging to achieve realistic transformations in geometric and high-level editing, where specific features are altered, and the overall consistency of the image needs to be preserved. 
\begin{figure}[t]
    \centering
    \includegraphics[width=0.9\linewidth]{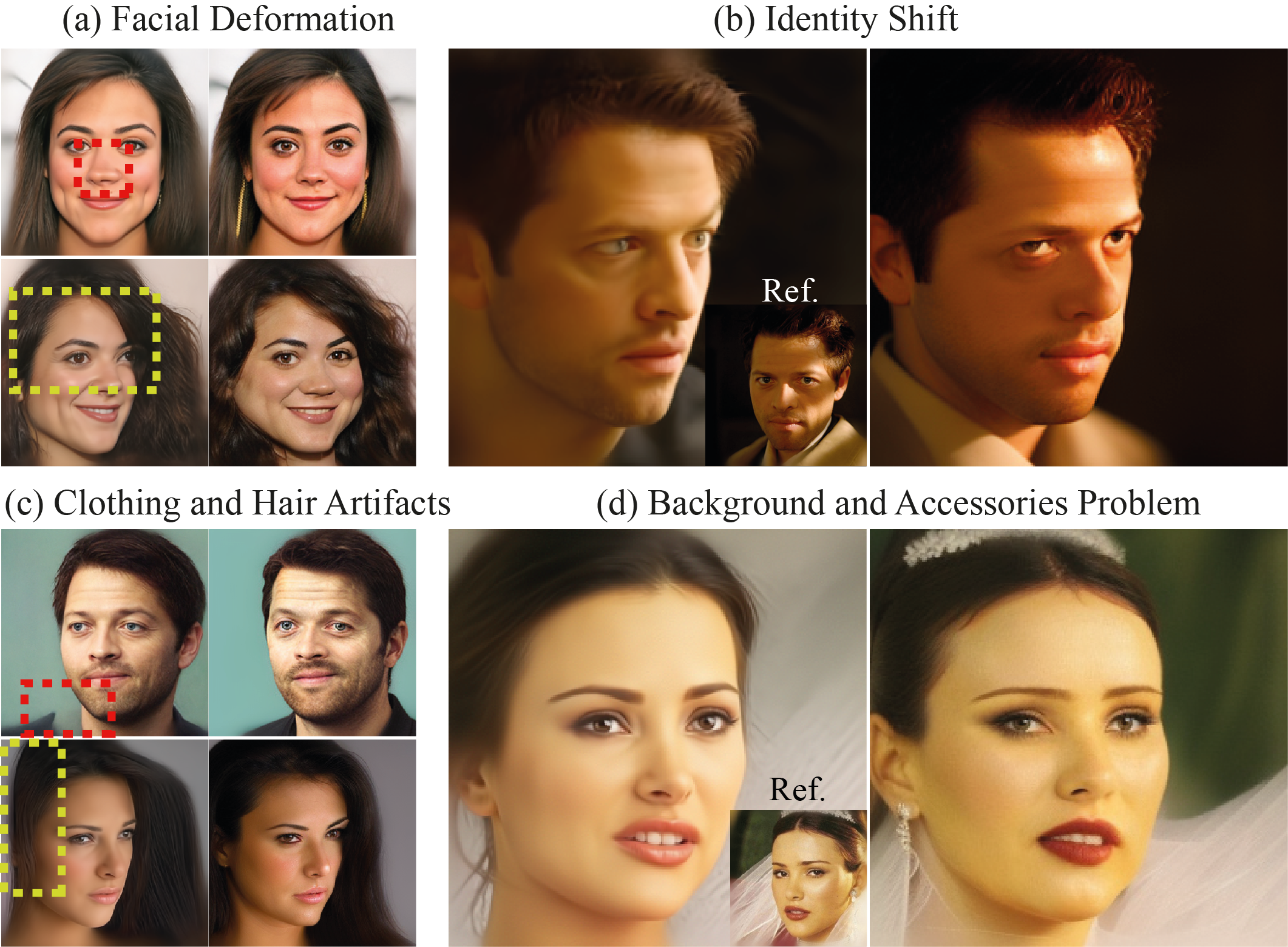} 
   \caption{Prior methods (left image of each pair) exhibit various types of issues, such as (a) unnatural facial deformations, (b) identity shifts in features like hair,  eye color, and face shape, (c) inconsistencies in clothing and hair styling, and (d) artifacts or distortions in the background and accessories. In contrast, our approach (right image of each pair) effectively resolves these issues, preserving natural facial geometry, consistent identity, and coherent styling across all elements. Reference images (Ref.) are provided for (b) and (d).
     }

    
    \vspace{-0.5cm} 
    \label{fig: challenges}
\end{figure}

The complexity further intensifies in facial image editing, where precise alterations in pose, expression, and lighting 
\begin{figure*}[t]
    \centering
    \includegraphics[width=0.9\textwidth, height=\textheight, keepaspectratio]{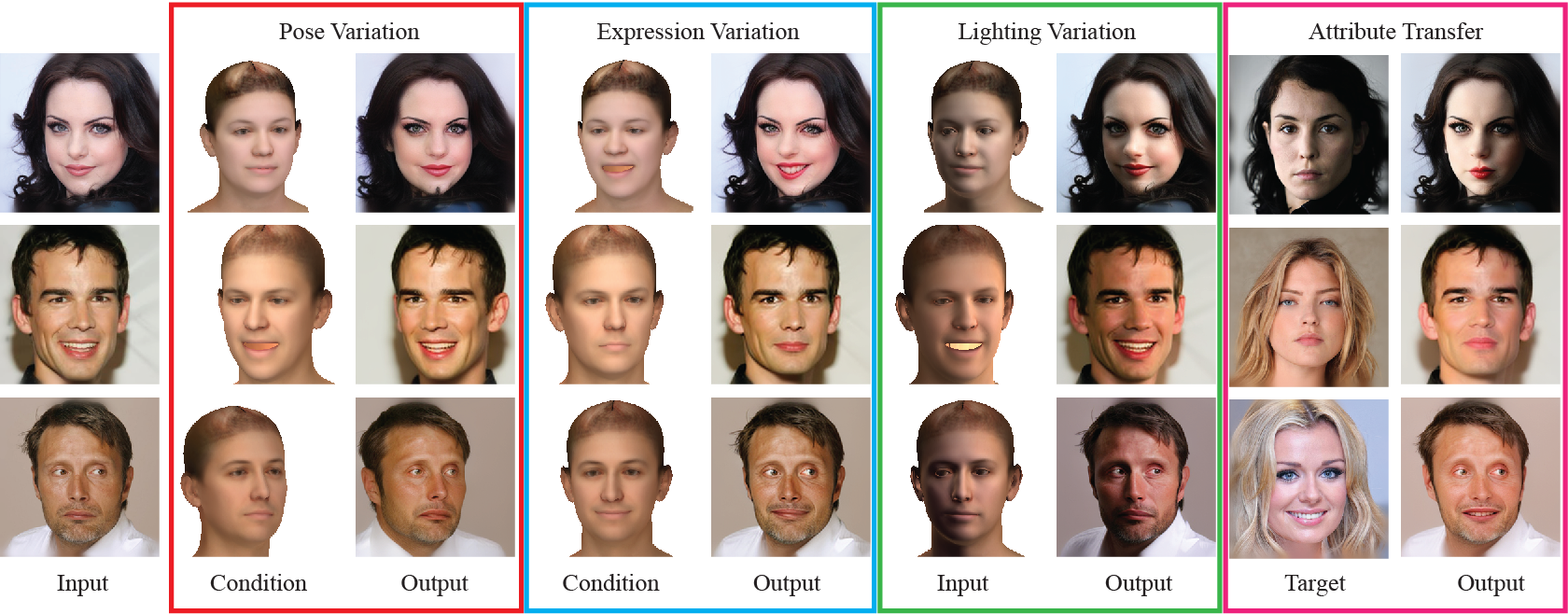}
    \caption{InstaFace leverages a single image to drive complex facial reenactments with conditional controls, including changes in pose, expression, and lighting. Our method ensures that the generated images retain the subject's identity, background, and fine-grained details while accurately reflecting the specified conditions.}
    \label{fig:teaser}
\end{figure*}
are desired while the individual`s identity needs to be preserved. Achieving precise and photorealistic facial image editing would open doors for various applications, such as personalized content creation~\cite{wang2023styleavatar}, digital avatars for gaming and virtual reality~\cite{xiang2024flashavatar}, and realistic interactions in virtual environments~\cite{salagean2023meeting}. 

Diffusion models~\cite{ho2020denoising} have demonstrated remarkable capabilities in image generation and manipulation. Recent works~\cite{jia2024discontrolface, preechakul2022diffusion} have adapted these models for more controllable editing of facial attributes. Among them
Animate Anyone~\cite{hu2024animate} adopted an approach similar to ControlNet~\cite{zhang2023adding}, where the identity reference image is introduced via a ReferenceNet, and the posing condition is introduced via a trainable pose guider. However, training this model requires multiple frames per identity, which limits its practicality for single-image editing. More importantly, using the same image for each identity will lead to overfitting, causing the model to copy the reference image while ignoring the intended control. Also, extending this approach to handle multiple conditions would require separate trainable modules for each of the conditions \cite{zhu2024champ}, significantly increasing training resource requirements. In order to achieve more accurate control over pose, expression, and lighting, DiffusionRig~\cite{ding2023diffusionrig} proposes to incorporate conditional maps from 3DMMs~\cite{blanz2023morphable, paysan20093d, luthi2017gaussian, gerig2018morphable} within a diffusion model. 
However, its reliance on multiple inference images limited its applicability, and its modification to pre-trained diffusion weights restricted its compatibility with open-source frameworks. More importantly, the facial feature and control conditions are not properly disentangled, leading to facial distortion and misalignment between facial features and non-facial features such as hair, neck, and accessories, as illustrated in Figure~\ref{fig: challenges}.


To address the issues of both paradigms, we introduce InstaFace, a novel approach that efficiently controls facial attributes while preserving identity with only a single inference image. Inspired by how Stable Diffusion effectively operates on the latent noisy input, we designed a 3D Fusion Controller Module for processing conditional maps in the latent space. We argue that if latent diffusion models can successfully use the latent space for input images, then it can be extended to conditional maps. This approach efficiently processes multiple conditional maps without requiring any additional trainable module, significantly reducing memory usage and computational overhead. This module then integrates with a Guidance Network, identical to the denoising UNet, which ensures that intended edits, such as pose, expression, and lighting, are preserved. By combining the latent representations from the guidance network with those from the diffusion network at each attention layer, our approach effectively utilizes both spatial and 3D information, therefore achieving precise control over desired facial attributes. 
Moreover, an Identity Preserver Module is introduced to better capture identity features and the overall semantic features. We propose integrating a face recognition encoder with a CLIP image encoder. The face recognition encoder focuses on preserving detailed facial features, while the CLIP encoder captures the broader semantic context, including elements like background and accessories.
Figure~\ref{fig:teaser} illustrates the results of the rigging achieved by our model with \textit{a single inference image} only.

We train our model using the FFHQ dataset ~\cite{karras2019style}, where conditional maps such as albedo maps, surface normal maps, and Lambertian render maps are extracted from the facial morphable model using the DECA~\cite{feng2021learning} to learn facial reenactment attributes. After learning these conditionals, we fine-tune our model with only a single inference image, where our newly introduced Identity Preserver Module effectively captures identity and semantic information, ensuring consistency. We provide extensive experimental results demonstrating superior performance compared to previous methods, along with ablation studies that highlight the improvements achieved at each stage of training and within each module. 
In summary, our contributions are as follows:
\begin{itemize}
\item We introduce a novel framework that more effectively incorporates the 3D conditional maps into an off-the-shelf diffusion model, achieving the new state-of-the-art identity-preserving facial attribute editing, with only a single inference image;
\item We introduce a novel Identity Preserver Module that combines a pre-trained multimodal vision model with a facial recognition model, ensuring maximal consistency of both the identity and the overall semantics in the input image;
\item We present comprehensive experimental results and ablation studies, demonstrating how our approach outperforms previous methods, and how each module contributes to enhancing control over facial attributes while maintaining identity consistency. 
\end{itemize}

\section{Related Work}
Our work is at the intersection of generative face models, 3D Morphable Face Models (3DMMs), and identity-preserving synthesis.

\textbf{Generative Facial Synthesis: }
Generative Adversarial Networks (GANs) have significantly advanced face generation, producing photorealistic images across various facial attributes \cite{hu2018pose, karras2020analyzing, tripathy2020icface}. However, these models struggle to disentangle and independently control attributes like appearance, shape, and expression, limiting their effectiveness in detailed editing. To address these issues approaches like \cite{marriott20213d, ghosh2020gif} incorporate 3D features for better attribute control. Diffusion models have emerged as the state-of-the-art in deep generative modeling, surpassing GANs in image synthesis \cite{dhariwal2021diffusion} and demonstrating their effectiveness in generating realistic facial images \cite{banerjee2023identity, valevski2023face0, xu2024personalized}. DisControlFace~\cite{jia2024discontrolface} leverages Diff-AE~\cite{preechakul2022diffusion}, using random masking techniques for effective training. However, these models struggle with diverse or large pose variations due to Diff-AE's reliance on near approximation. DiffusionRig~\cite{ding2023diffusionrig} enhances synthesis with pixel-aligned conditions (e.g., normals, albedo) and uses multiple images for identity preservation but still faces challenges in maintaining consistency across generated outputs. CapHuman~\cite{liang2024caphuman} uses textual data to control facial attributes but struggles with consistency, leading to variations in background, hair, and facial shape. VOODOO 3D~\cite{tran2024voodoo} addresses volumetric head reenactment but struggles with pose control, leading to unnatural tilts of the entire input image and visual artifacts.

Our method, InstaFace, uniquely addresses these limitations by retaining identity with just one image, even under large variations in conditionals, using a combination of CLIP and a face recognition model.

\textbf{Condition-Driven Face Synthesis: }
Effective facial synthesis and editing generally rely on integrating conditional inputs to guide the generation process. For instance, GANs, particularly StyleGAN, excel in transferring styles from a constant input tensor (4×4×512) to produce high-fidelity images by feeding latent code \( z \in \mathbb{Z} \) through different routes to the network. Similarly, diffusion models, like DDPMs~\cite{ho2020denoising}, use text embeddings from large pre-trained models or employ encoders to generate latent codes that guide the noise prediction and denoising processes. For facial image editing, where retaining the original image features while altering global attributes such as pose and lighting or local attributes like expressions (mainly mouth, eyes, and cheeks) is essential, incorporating 3D perspectives becomes necessary. Methods such as ~\cite{ding2023diffusionrig, jia2024discontrolface, ghosh2020gif} achieve this by utilizing albedo maps, normal maps, and Lambertian renders from 3DMM models ~\cite{li2017learning, ichim2017phace, koppen2018gaussian} to condition their generative models. To utilize these conditionals effectively, ControlNet~\cite{zhang2023adding} stands out for its ability to guide the denoising UNet spatially, layer by layer, due to its similar structure to the denoising UNet. However, previous methods like DisControlFace face challenges with the availability of pre-trained models. While many pre-trained Stable Diffusion models exist due to their generative capability on large datasets, the diverse nature of conditionals means there can be various types of ControlNet, complicating the use of pre-trained models for ControlNet. In our approach, we leverage the same Stable Diffusion structure, ensuring ease of training and effective integration of conditionals, thereby overcoming these challenges.

\begin{figure*}[t]
    \centering
    \includegraphics[width=0.9\textwidth, height=\textheight, keepaspectratio]{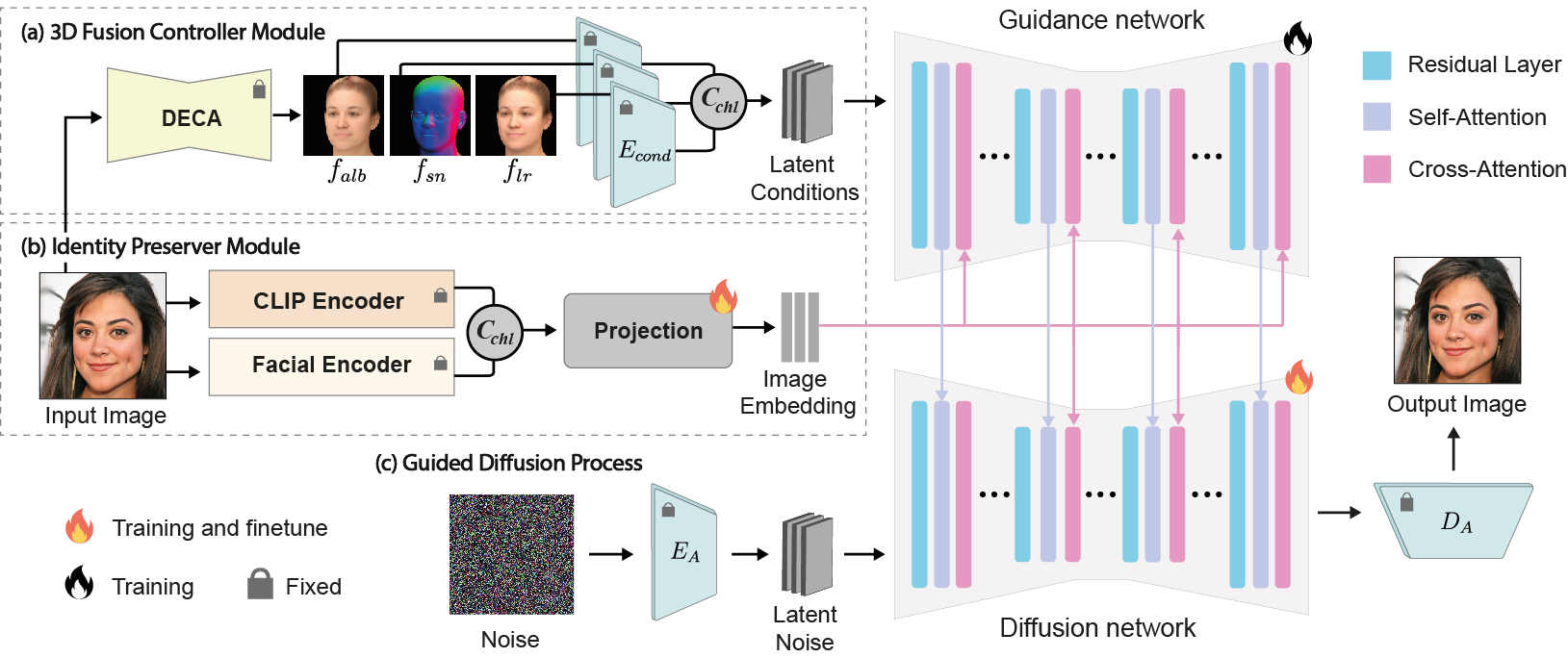}
    \caption{\textbf{Overview of InstaFace Architecture:}
(a) Conditional maps generated by the pre-trained DECA Model are processed by the 3D Fusion Controller to produce latent conditionals, which are then utilized by the Guidance Network to guide the diffusion model;
(b) Semantic and identity features are extracted and concatenated to provide conditions for the diffusion process;
(c) The Diffusion Network synthesizes the final image, guided by both the Guidance Network and the concatenated embeddings.}
    \label{fig:Architecture}
\end{figure*}

\section{Preliminaries}
In this section, we provide the foundational knowledge of 3D morphble models (3DMM) and stable diffusion (SD), which are essential for our method.

\textbf{3D Morphable Face Models:}
We use FLAME as the 3D Morphable Model (3DMM), which leverages linear blend skinning (LBS) with pose-dependent corrective blendshapes to represent head pose, face geometry, and facial expressions. 
The FLAME model is defined by $M(\beta, \theta, \psi)$, where the template mesh,
$$T_P(\beta, \theta, \psi) = \bar{T} + B_S(\beta; S) + B_P(\theta; P) + B_E(\psi; E), $$ 
combines shape $\beta$, pose $\theta$, and expression $\psi$ parameters to create a detailed facial representation. Here, $\bar{T}$ is the mean template in canonical pose, $B_S$ denote the shape blendshapes, $B_P$ denote the pose blendshapes, and $ B_E$ denote the expression blendshapes. 
To complement FLAME, DECA enhances the 3DMM by integrating an appearance model that predicts detailed facial geometry, albedo, and lighting from single in-the-wild images. Specifically, DECA encodes 2D images into FLAME parameters,~\ie, $\theta$, $\psi$, and $\beta$, along with lighting $l$ and camera $c$ settilgs and captures facial attributes via generating a displacement map. After decoding, it generates albedo maps, surface normals, and spherical harmonic (SH) lighting. These maps are then used to guide the diffusion model, transferring the DECA-predicted face to a photorealistic image while maintaining detailed facial attributes.

\textbf{Stable Diffusion:}
Our method is built upon Stable Diffusion ~\cite{rombach2022high}, which performs the diffusion process efficiently in the latent space rather than the pixel space.  It consists of an encoder, $E$, which maps an input image, $x$, into a latent representation, $z = E(x)$. Stable Diffusion utilizes this latent representation to perform the diffusion and denoising processes. During training, the latent representation $z$ is iteratively diffused over $t$ timesteps, generating noisy latents, $z_t$, given by
$$z_t = \sqrt{\bar{\alpha}_t} z_0 + \sqrt{1 - \bar{\alpha}_t} \epsilon,$$
which are then denoised by a UNet ~\cite{ronneberger2015u} to predict the original latent representation. The training objective for Stable Diffusion, denoted as $\epsilon_\theta$, aims to predict noise $\epsilon \sim \mathcal{N}(0, I)$. The objective is expressed as follows:
$$L_{\text{simple}} = \mathbb{E}_{z_0, \epsilon \sim \mathcal{N}(0, I), c, t} \left[ \left\| \epsilon - \epsilon_\theta (z_t, c, t) \right\|^2 \right],$$
where $z_0$ represents the original latent code, $t$ is the time step within the diffusion process, and the predefined functions $\bar{\alpha}_t$ govern the progression of noise during the diffusion process. 
$c$ incorporates additional conditional information to steer the denoising process.
During inference, the process begins with sampling $z_T$ from a Gaussian distribution, then progressively denoised to $z_0$ using a deterministic sampling process, such as DDPM~\cite{ho2020denoising} or DDIM~\cite{song2020denoising}. In each step, the denoising UNet predicts the noise for the corresponding timestep $t$. Finally, the decoder $D$ reconstructs $z_0$ back into the image space, yielding the final image.


\section{Methodology}

To achieve robust facial image editing, we propose a generative framework,~\ie, \textbf{InstaFace}, as illustrated in Figure~\ref{fig:Architecture}. Our method integrates control conditions from  3D Morphable Model (3DMM) to edit the target face, accordingly. InstaFace is composed of two stages. In the first stage (Figure~\ref{fig:Architecture}a), the Guidance network learns to understand and generalize facial attributes from a broad dataset of reference images. This ensures that the generative process can effectively handle various appearance conditions, such as pose, expression, and lighting. In the second stage (Figure~\ref{fig:Architecture}b), InstaFace fine-tunes its generative capabilities using a specific target image, referred to as the inference image. During this stage, the embeddings obtained from pre-trained CLIP and facial recognition models provide the necessary information to guide the diffusion process. This guidance is crucial for accurately retaining the specific identity of the individual, ensuring that facial features and background attributes, such as hair and accessories, are preserved. Finally, as shown in Figure~\ref{fig:Architecture}c, the Diffusion Network learns both the overall identity information and the necessary conditioning to generate the desired facial image. This combination allows InstaFace to deliver high-fidelity, identity-preserving facial edits with precise attribute modifications.



\subsection{Facial Condition Adaptation}

Our initial stage aims to learn facial priors from 3DMM-generated conditionals, focusing on identifying specific editable attributes and capturing high-level features. To achieve this, we employ DECA ($E_{\text{DECA}}$) to estimate FLAME parameters from 2D images, effectively bridging the image data to the 3D domain for detailed facial representation. Specifically, DECA predicts the shape $\beta$, pose $\theta$, and expression $\psi$ parameters, along with the lighting $l$ and camera $c$ settings, using the FLAME model with its appearance and illumination models. These parameters are then used to generate pixel-aligned maps, including albedo maps ($f_{alb}$), surface normals, and Lambertian renderings, with the help of the DECA decoder ($D_{\text{DECA}}$). By translating non-spatial 3DMM parameters into spatially meaningful visual representations, we ensure that the model can accurately capture the detailed geometry and appearance of the face.

To efficiently utilize these 3D features and condition the whole model, we introduce the 3D Fusion Controller—a core contribution of our approach. The 3D Fusion Controller takes the 3DMM-generated conditionals—such as albedo maps, surface normals, and rendered maps—and converts them into latent space representations using a pre-trained frozen autoencoder ($E_{\text{cond}}$). These latent conditionals are then concatenated in the channel dimension, allowing us to handle multiple conditionals simultaneously without introducing additional trainable parameters. This simple yet effective design enables our model to retain and leverage the latent 3D conditions more efficiently, leading to a robust conditioning process without additional computational overhead.

Now, instead of directly passing the latent conditional maps to the main diffusion process, we employ a ControlNet-like architecture, which we refer to as the Guidance Network ($\mathcal{G}$). This allows us to retain the fundamental weights of the Diffusion Network ($\mathcal{M}_{\text{diff}}$) unchanged, providing efficiency without compromising the model's initial learned capabilities. Additionally, this ControlNet-like approach leverages the robust feature extraction capabilities of U-Net architectures, enabling spatial-aware conditioning for the diffusion process. Inspired by~\cite{hu2024animate}, we structured the Guidance Network to mirror the denoising U-Net architecture within our framework. The Guidance Network benefits from pre-trained image feature modeling capabilities by inheriting weights from the original Stable Diffusion model, ensuring a well-initialized feature space. This avoids the need to train from scratch or rely on existing ControlNet models, which are not suitable for our 3D facial image editing task. The Guidance Network processes the latent conditional maps generated by the 3D Fusion Controller, and from each layer, the conditioning information flows to the Main Diffusion Network.

The main input image is also encoded into a latent representation using a pre-trained autoencoder ($E_A$). Noise is added to these latent representations using DDIM and then passed into the Diffusion Network ($\mathcal{M}_{\text{diff}}$). Consequently, the intermediate features of the Guidance Network are spatially combined with the corresponding intermediate features of the Diffusion Network in the attention module, specifically just before the self-attention layers. This integration ensures that the conditioning information from the Guidance Network effectively influences the noise prediction process.


Formally, the process can be described by the following equation:
\begin{align}
F_{\text{comb}} ={}& \mathcal{M}_{\text{diff}}(E_A(x)) \nonumber \\
&\oplus \mathcal{G} \left( \text{concat}_{\text{chl}} \left( E_{cond} (D_{\text{DECA}} \left( E_{\text{DECA}}(x) \right) \right) \right)) \label{eq:combined},
\end{align}
where $\oplus$ indicates that the features from the Guidance Network are added to the intermediate noisy feature maps of the Diffusion Network before passing through the self-attention layers.

\subsection{Identity Preserving Guidance}
The Diffusion Network starts from complete noise during inference, which is why it is crucial how the guidance is provided for the main input image. Unlike style-editing methods (e.g., transforming images into paintings, sketches, or anime) \cite{wang2024instantid}, our specific task requires maintaining the identity of the given input person and the background or accessories while allowing changes in pose, expression, and lighting. Specifically, in text-to-image tasks, high-level semantics suffice, but image-based generation demands detailed guidance to preserve both identity and fine-grained attributes. In this case, CLIP excels at capturing high-level semantic information and contextual understanding from images, which is beneficial for generating coherent and contextually accurate outputs \cite{song2023objectstitch, yang2023paint, ye2023ip}. However, CLIP’s limitation lies in its reliance on low-resolution images during encoding, which results in the loss of fine-grained details crucial for high-fidelity image synthesis. Additionally, CLIP's training primarily focuses on matching semantic features for text-image pairs, which may lead to insufficient encoding of detailed facial attributes and unique identity features \cite{hu2024animate}.  This issue is compounded by the fact that CLIP is trained on weakly aligned datasets, which tends to emphasize only broad attributes such as layout, aesthetic, and color schemes \cite{wang2024instantid}.

In contrast, facial recognition technology has seen remarkable advancements in computer vision systems, demonstrating exceptional accuracy in identifying individuals. Leveraging a facial recognition model can be an effective approach to capture and retain fine-grained identity details, ensuring the generated images preserve the unique attributes of the input face. However, relying solely on facial recognition models can pose challenges. These models often generate embeddings that focus primarily on specific facial regions, such as the eyes, cheeks, and nose \cite{zee2019enhancing}. This selective focus may lead to inconsistencies in other parts of the face, resulting in unrealistic image synthesis. 


To address these limitations, we propose a novel approach that combines the strengths of CLIP and facial recognition models. Specifically, we combine the image embeddings from the CLIP model $E_{\text{CLIP}}$ with the detailed identity embeddings generated by a face recognition model $E_{\text{FR}}$. 
These combined embeddings are processed through a projection module, which incorporates a series of attention mechanisms and feedforward networks. The projected embedding is then used in the cross-attention mechanism of the Guidance and Diffusion networks. This dual-embedding strategy ensures that the generated images retain high-level semantic coherence from CLIP while capturing fine-grained identity details from the face recognition model, thus overcoming the shortcomings of using either model independently.

The combined embedding \( E_{\text{comb}} \) is computed as follows:
\begin{align}
E_{\text{comb}} = \text{Proj}(E_{\text{CLIP}}(x), E_{\text{FR}}(x)),
\label{eq:combined_embedding}
\end{align}
where \(\text{Proj}\) denotes the projection module that merges the feature embeddings using attention and feedforward layers. This combined embedding is then incorporated into the cross-attention mechanisms of both the Guidance Network and the main Diffusion Network.

Therefore, the overall loss function for training our model is defined as:
\begin{align}
\mathcal{L} = \mathbb{E}_{z_0, t, E_{\text{comb}}, c_f, \epsilon \sim \mathcal{N}(0, 1)} \left[ \left\| \epsilon - \epsilon_\theta \left( z_t, t, E_{\text{comb}}, c_f \right) \right\|^2 \right],
\label{eq:combined_2}
\end{align}
where $c_f$ represents the layer-wise features extracted from the guidance network $\mathcal{G}$.

\begin{figure*}[t]
    \centering
    \includegraphics[width=0.90\textwidth]{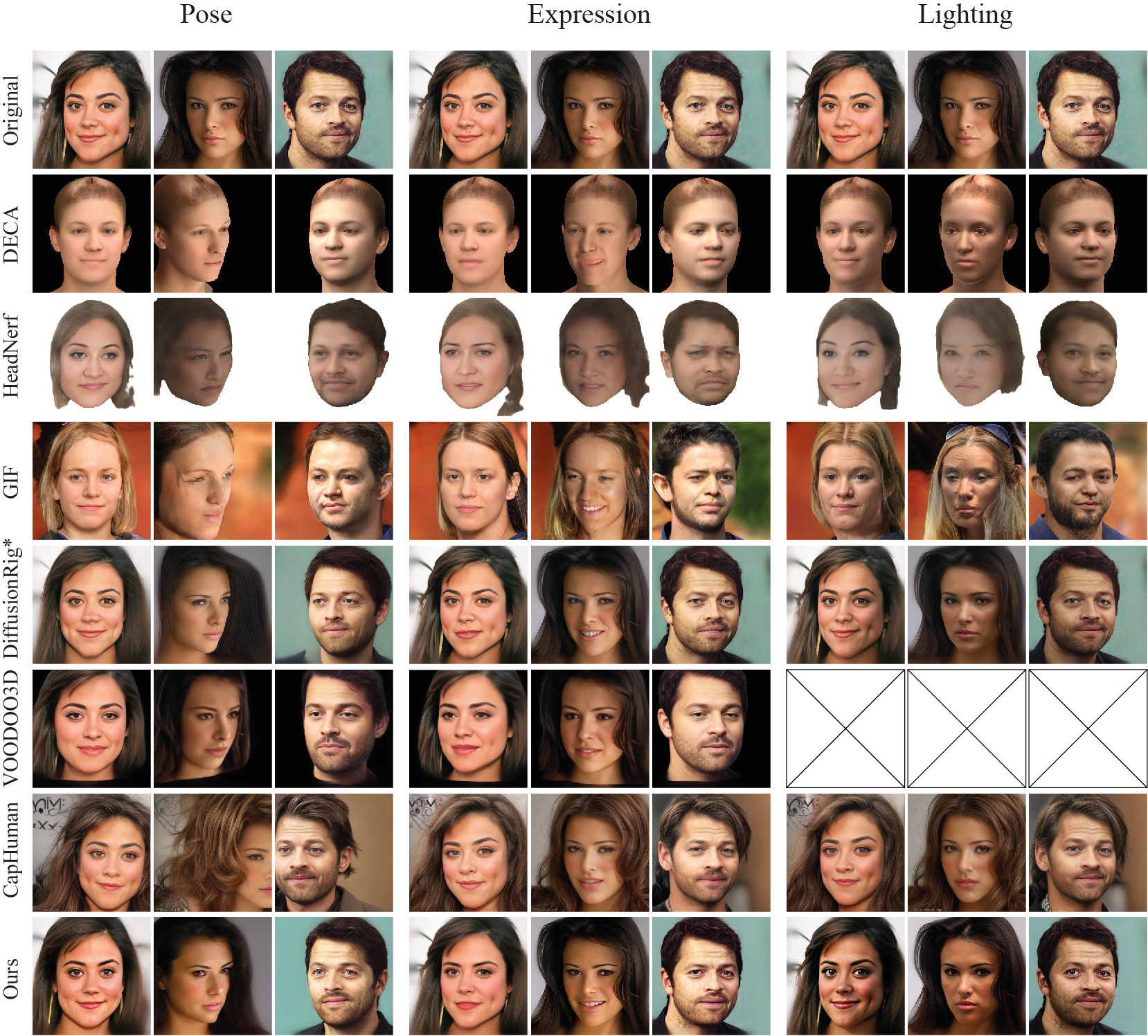}
    \caption{Baseline comparisons with DECA, HeadNerf, GIF, DiffusionRig, CapHuman, and VOODOO3D. Our method performs better in retaining identity while generating realistic facial images under varying conditions. Here, DiffusionRig is marked with (*) as it necessitates per-subject fine-tuning using a set of 20 images. VOODOO3D does not support lighting variation edits.}
    \label{fig:comparison}
\end{figure*}

\subsection{Training Strategy}
The training process is divided into two stages. In the first stage, our model learns conditional attributes and general facial features. We initialize training using pre-trained weights from Stable Diffusion (SD) for both the Guidance Network and the Main Diffusion Network. The 3D pixel-aligned conditionals, generated from reference input images using the pre-trained DECA model, are processed by the 3D Fusion Controller before being passed to the Guidance Network. Corresponding features from the Guidance Network are integrated into the Diffusion Network before each self-attention layer. The reference input image is processed through a VAE, a face recognition model, and CLIP. The latents from the VAE go to the Main Diffusion Network, while the projected embeddings from the Identity Preserver Module, derived from CLIP and the face recognition model, are fed into the cross-attention mechanisms of both the Guidance Network and the Diffusion Network via the projection module. During this stage, the Guidance Network, Main Diffusion Network, and projection module are trainable, while the VAE, CLIP, and face recognition models are kept fixed. In the second stage, we fine-tune our model using only the inference image, which serves as the target image for inference. Only the projection module and the Diffusion Network are trainable in this stage, while all other components retain their learned weights. This approach enables the model to adapt to specific facial attributes, enhancing both accuracy and realism in the generated outputs.

\section{Experiments}
\subsection{Implemetation }

In the first stage of our training, we utilize the FFHQ dataset ~\cite{karras2019style}, which contains 70,000 high-quality facial images. The images are first resized to 256x256 pixels to match the input requirements of the VAE encoder. After processing through the VAE, the images are converted into latent representations with a size of 32x32 and 4 channels. We conduct our experiments on 2 NVIDIA Quadro RTX 8000 GPUs, each with a batch size of 6, for a total of 55,000 steps. 
The learning rate is set to 1e-5, and we use the AdamW optimizer during this stage. In the second stage, we fine-tune the model using a single inference image to retain the identity. We create copies of this image to form a batch size of 8 and trained the model for 50 steps, which has empirically provided the best results. During this phase, the learning rate remains at 1e-5, and we continue to use the AdamW optimizer. During inference, we can either specify FLAME parameters for DECA to generate the required conditional maps (first 3 columns of Figure~\ref{fig:teaser} or use another image, referred to as the target image from which DECA extracts these maps (last column of Figure~\ref{fig:teaser}). We then employ the DDIM ~\cite{song2020denoising} sampler with 20 denoising steps.



\subsection{Comparisons}
To assess the efficacy of our approach, we perform comparisons with cutting-edge techniques such as HeadNerf\cite{hong2022headnerf}, GIF\cite{ghosh2020gif}, DiffusionRig\cite{ding2023diffusionrig}, CapHuman~\cite{liang2024caphuman}, and VOODOO3D~\cite{tran2024voodoo} as depicted in Figure~\ref{fig:comparison}. Our approach consistently surpasses these reference points in producing authentic facial photographs while preserving identity. GIF efficiently alters facial attributes but struggles with identity preservation. HeadNerf captures facial identification well but fails to maintain structural integrity when the face turns away from the frontal view. While DiffusionRig performs well, it produces artifacts when adjusting pose due to remnants of the original image. VOODOO3D tilts the entire image instead of following the driver's pose and cannot handle lighting edits. CapHuman struggles with retaining background accessories and hair and has issues with camera shifting. In this case, Our approach stands out because we utilize separate conditional inputs and keep the ControlNet fixed during fine-tuning. This results in the generation of authentic images that sustain both the distinguishing characteristics and the desired traits.

\textbf{Image Quality and Identity Evaluation.}
To facilitate a quantitative comparison, we adopt the same experimental setup as DiffusionRig, generating a set of 400 images, each with a unique pose and expression
We evaluate the images using facial reidentification models(Re-ID) \cite{king2009dlib}, Perceptual Similarity (LPIPS) \cite{zhang2018unreasonable}, Frechet Inception Distance (FID) \cite{heusel2017gans}, and Structural Similarity Index (SSIM) \cite{wang2004image}. To provide a basis for comparison, we incorporate findings from DiffusionRig and CapHuman, which are, in this case, most relatable to our method. Nevertheless, our model consistently outperforms previous methods in almost all measurements, as demonstrated in Table~\ref{tab:quant_eval}.

\FloatBarrier
\begin{table*}[!htbp]
\centering
\caption{Quantitative evaluation for novel pose, expression, and lighting synthesis.}
\setlength{\tabcolsep}{3pt} 
\renewcommand{\arraystretch}{1.1} 
\tiny 
\resizebox{\textwidth}{!}{
\begin{tabular}{lcccc|cccc|cccc}
    \toprule
    & \multicolumn{4}{c|}{Novel Pose Synthesis} & \multicolumn{4}{c|}{Novel Expression Synthesis} & \multicolumn{4}{c}{Novel Lighting Synthesis} \\
    \cmidrule(r){2-5} \cmidrule(r){6-9} \cmidrule(l){10-13}
    Method & LPIPS$\downarrow$ & SSIM$\uparrow$ & FID$\downarrow$ & Re-ID$\uparrow$ & LPIPS$\downarrow$ & SSIM$\uparrow$ & FID$\downarrow$ & Re-ID$\uparrow$ & LPIPS$\downarrow$ & SSIM$\uparrow$ & FID$\downarrow$ & Re-ID$\uparrow$ \\
    \midrule
    CapHuman & 0.5914 & 0.3968 & 205.25 & 95.63 & 0.4762 & 0.4487 & 194.00 & 95.57 & 0.4772 & 0.4431 & 186.85 & 95.86 \\
    DiffusionRig* & 0.4547 & 0.4222 & 73.16 & 96.72 & 0.1891 & 0.6625 & 68.20 & 97.27 & 0.2367 & 0.5788 & 69.28 & \textbf{97.27} \\
    Ours & \textbf{0.3747} & \textbf{0.6160} & \textbf{53.63} & \textbf{96.81} & \textbf{0.0931} & \textbf{0.8860} & \textbf{46.58} & \textbf{97.40} & \textbf{0.1436} & \textbf{0.8033} & \textbf{43.72} & 97.22 \\
    \bottomrule
\end{tabular}}
\label{tab:quant_eval}
\end{table*}

\begin{figure}[ht]
    \centering
    \includegraphics[width=0.95\linewidth]{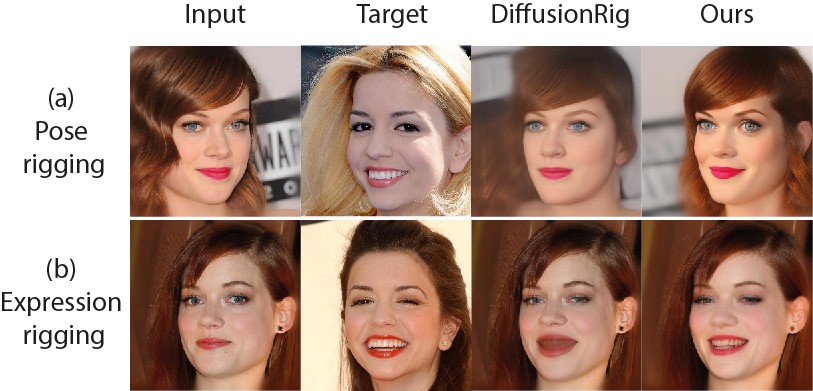} 

\caption{Evaluation of pose and expression rigging quality during single-image fine-tuning.The input images are expected to follow the corresponding target image's (a) pose and (b) expression.}
    \label{fig:pose_rigging_quality}
\end{figure}

\textbf{Rigging Quality Evaluation.}
In this experiment, we evaluate the rigging quality of our model by generating 1200 images to assess how accurately it conforms to the desired pose, expression, and shape. Unlike prior studies, we do not randomly select images for evaluation. To ensure a fair comparison, we include CapHuman~\cite{liang2024caphuman} and conduct a single-image inference for both DiffusionRig\cite{ding2023diffusionrig} and our model, using the same parameters outlined in DiffusionRig's paper. This allows us to assess the effectiveness of our single-image fine-tuning in maintaining control and quality. We then measure the DECA~\cite{li2017learning}re-inference error to compare the results. The evaluation results, as shown in Table~\ref{tab:deca_reinference}, our model shows improved pose control, achieving an error of only 9.37 mm compared to DiffusionRig and CapHuman. The key reason for this improvement is our model's approach to handling control inputs. Unlike DiffusionRig, which merges conditional maps with the reference image, which leads to distortions during single-image fine-tuning, our model keeps these maps separate. This disentanglement ensures precise pose control, as illustrated in Figure~\ref{fig:pose_rigging_quality}a. DiffusionRig achieves better expression accuracy with an error of 3.37 mm, compared to our model's 5.14 mm, while CapHuman has a higher error of 7.37 mm. However, DiffusionRig often produces artifacts around the mouth area~\ref{fig:pose_rigging_quality}b, resulting from its attempt to maintain pixel consistency from the reference image. While slightly less accurate for expression, our approach avoids these artifacts, resulting in cleaner and more natural outputs.

\begin{table}[!htbp]\vspace{-0.1cm} 
\centering
\caption{DECA re-inference error evaluation based on facial-landmarks}\vspace{-0.1cm} 
\setlength{\tabcolsep}{10pt} 
\renewcommand{\arraystretch}{1.1} 
\begin{tabular}{lcc}
    \toprule
    Method & Pose & Expression \\
    \midrule
    CapHuman & 23.51 mm & 7.37 mm \\
    DiffusionRig & 11.32 mm & \textbf{3.37 mm} \\
    Ours & \textbf{9.37 mm} & 5.14 mm\\
    \bottomrule
\end{tabular}
\vspace{-0.4cm} 
\label{tab:deca_reinference}
\end{table}

\vspace{-0.4cm}
\subsection{Ablation Studies}
We conduct an ablation study to showcase the efficacy of integrating CLIP with a facial recognition system, generating 600 photos for quantitative analysis.


\textbf{Impact of Facial Recognizer.} The facial recognizer (FR) is designed to capture critical identity-specific features. To assess its impact, we conduct ablation studies with and without the FR using the Re-identification (Re-ID) metric \cite{king2009dlib} to measure identity consistency across pose, expression, and lighting variations. As shown in Table~\ref{tab:facial_recognizer_ablation}, while the improvement may appear minor, it is crucial, as the FR plays a significant role in preserving essential facial features, such as the eyes, nose, and mouth, which are key to maintaining the subject's identity.







\textbf{Impact of CLIP Encoder.} Our approach integrates the CLIP encoder to provide an accurate representation of the input image and maintain its original distribution. To evaluate its effectiveness, we use the Fréchet Inception Distance (FID) \cite{heusel2017gans} and Structural Similarity Index Measure (SSIM) \cite{wang2004image} to quantify image realism and fidelity. As shown in Table~\ref{tab:clip_encoder_ablation}, the CLIP encoder significantly improves fidelity, especially for pose variations, while also preserving important background details such as the neck, hair, and accessories. 

\vspace{-0.1cm} 
\begin{table}[!htbp]
    \centering
    \caption{Quantitative evaluation for ablation studies: Identity retention, image fidelity, and realism.}
    \setlength{\tabcolsep}{6pt} 
    \renewcommand{\arraystretch}{1.1} 
    \small 

    \begin{subtable}[t]{\columnwidth}
        \centering
        \caption{ Facial Recognizer Ablation}
        \label{tab:facial_recognizer_ablation}
        \begin{tabular*}{\columnwidth}{@{\extracolsep{\fill}}lccc@{}}
            \toprule
            Method & Pose & Expression & Lighting \\
            \midrule
            \multicolumn{4}{@{}l}{\textbf{\footnotesize Re-identification Accuracy (Re-ID\(\uparrow\))}} \\
            Base + CLIP & 96.20 & 97.00 & 96.96 \\
            Base + CLIP + FR & \textbf{96.72} & \textbf{97.28} & \textbf{97.12} \\
            \bottomrule
        \end{tabular*}
    \end{subtable}
    
    \vspace{0.2cm} 

    \begin{subtable}[t]{\columnwidth}
        \centering
        \caption{ CLIP Encoder Ablation}
        \label{tab:clip_encoder_ablation}
        \begin{tabular*}{\columnwidth}{@{\extracolsep{\fill}}lccc@{}}
            \toprule
            Method & Pose & Expression & Lighting \\
            \midrule
            \multicolumn{4}{@{}l}{\textbf{\footnotesize Fr\'echet Inception Distance (FID \(\downarrow\))}} \\
            Base + FR & 70.435 & 56.84 & 53.133 \\
            Base + FR + CLIP & \textbf{58.20} & \textbf{46.685} & \textbf{44.66} \\
            \midrule
            \multicolumn{4}{@{}l}{\textbf{\footnotesize Structural Similarity Index Measure (SSIM \(\uparrow\))}} \\
            Base + FR & 0.5766 & 0.8730 & 0.7660 \\
            Base + FR + CLIP & \textbf{0.5879} & \textbf{0.8920} & \textbf{0.7850} \\
            \bottomrule
        \end{tabular*}
    \end{subtable}

    \vspace{-0.5cm} 
\end{table}

\section{Limitations and Conclusion}
Despite achieving superior results compared to previous state-of-the-art methods, our method has limitations. Specifically, when the input image is in a cornered pose, the result can sometimes deviate from the desired identity. Additionally, issues with color accuracy can arise under extreme brightness conditions. As our approach utilizes the DECA model, minor deviations in expression can occur due to its estimation limits. 

In this paper, we developed an efficient approach for facial image editing using a novel architecture enhanced by the 3D Fusion Controller, Guidance Network, and Identity Preserver Module. Our method excels in retaining identity and accessories while allowing for fine-tuning with minimal steps. This results in high-fidelity images with accurate attribute editing, demonstrating significant advancements over previous \mbox{methods}.


\end{document}